\pdfoutput=1
\documentclass[10pt]{article}
\usepackage[letterpaper]{geometry}
\usepackage{hicss}
\usepackage{times}
\usepackage[none]{hyphenat}
\usepackage{url}
\usepackage{latexsym}
\usepackage{graphicx}
\graphicspath{{Fig/}}
\usepackage[T1]{fontenc}
\usepackage[utf8]{inputenc}
\usepackage{geometry}
\usepackage{fancyhdr}
\usepackage{amsfonts}
\usepackage{xpatch}
\usepackage{amsmath}
\usepackage{amsthm}
\usepackage{booktabs}
\usepackage{algorithm}
\usepackage{algorithmic}
\usepackage{dsfont}
\usepackage{array,booktabs}
\newcolumntype{L}{@{}>{\kern\tabcolsep}l<{\kern\tabcolsep}}
\usepackage{colortbl}
\usepackage{xcolor}
\usepackage{hyperref}
\title{A Novel Deep Learning Model for Hotel Demand and Revenue Prediction amid COVID-19}
\author{Ashkan Farhangi \\
 University of Central Florida \\
 {\underline{ashkan.farhangi@ucf.edu}} \\\And
 Arthur Huang \\
  University of Central Florida \\
 {\underline{arthur.huang@ucf.edu} }\\\And 
 Zhishan Guo \\
  University of Central Florida \\
 {\underline{zhishan.guo@ucf.edu}} \\}
\date{}
\begin{document}
\maketitle
\begin{abstract}
The COVID-19 pandemic has cast a substantial impact on the tourism and hospitality sector. Public policies such as travel restrictions and stay-at-home orders had significantly affected tourist activities and service businesses' operations and profitability. To this end, it is essential to develop an interpretable forecast model that supports managerial and organizational decision-making. We developed DemandNet, a novel deep learning framework for predicting time series data under the influence of the COVID-19 pandemic. The framework starts by selecting the top static and dynamic features embedded in the time series data. Then, it includes a nonlinear model which can provide interpretable insight into the previously seen data. Lastly, a prediction model is developed to leverage the above characteristics to make robust long-term forecasts. We evaluated the framework using daily hotel demand and revenue data from eight cities in the US. Our findings reveal that DemandNet outperforms the state-of-art models and can accurately predict the impact of the COVID-19 pandemic on hotel demand and revenues\footnote{Our code and datasets are available at \texttt{https://github.com/ashfarhangi/COVID-19}}.
\end{abstract}
\section{Introduction}
The COVID-19 pandemic has substantially disrupted global economic activities. The stay-at-home order and business closure policy intended to curb the COVID-19 transmission have taken an unintended toll on the economy \cite{introeconomysurvey}. As large volumes of business data become available, it is critical to develop machine learning models and tools to shed light on the impact of COVID-19 and policies on demand and revenue and unpack the future trends \cite{introNaturePoliciesfuture, experimentDataSpendingshousehold}. Such machine learning models should offer both high accuracy and interpretability to assist managers in making informed decisions about new business strategies and market recovery. 
The fine-grained business data amid crises are high-dimensional time series observations with great complexity. They are difficult to be sufficiently modeled using traditional time series models due to their dynamic nature and not frequently seen uncertainties~\cite{ methnNNmodelingreview, ArimaOG, farhangi}. Furthermore, current deep learning models contain complex and nonlinear parameters that are often uninterpretable~\cite{lim2019temporal}. In addition, they rarely provide uncertainty measurement in their predictions, which becomes a challenge when the data are influenced by external shocks such as the COVID-19 pandemic \cite{leader}. 

Given that the COVID-19 pandemic has caused a sudden change in the data trend and seasonality, it is vital to develop a new deep learning approach for modeling real-time data under the influence of external shocks. The new approach should be interpretable and reliable and would uncover the effects (e.g., correlation and strengths) of exogenous variables on the interested variables.

To this end, we propose DemandNet, a novel deep learning framework that can capture the essential features in time series data and achieve interpretability. Specifically, a carefully designed feature selection mechanism is proposed to filter appropriate information from the prior pandemic data. Furthermore, we develop a multilayer neural network to derive the nonlinear relation of the selected features to specific economic responses (business and individuals). The new approach is tested on daily hotel demand and revenue data in eight major cities in the US between 2013 and 2020. The results show that the model is able to learn the impact of COVID-19 from the training set and make accurate predictions of hotel demand and revenue on unseen (not trained) time series data. 
\vspace{1mm}
\noindent{\bf Contributions.} 
This study includes the following contributions:
\begin{itemize}
 \item We design a feature selection mechanism to select the top static and dynamic features of a time series. Such capability also enhances the DemandNet's ability to capture complex critical features.
 \item We design a multilayer neural network to derive the nonlinear relation of the selected features to the predictor. Such nonlinear relation can then provide interpretable insights about the past and be an aid in future prediction.
 \item We developed a novel multi-step time series prediction model that leverages a dynamic dropout optimization mechanism to minimize the prediction uncertainties and provide an optimal level of confidence in forecasts. 
 \item The framework is capable of predicting newly added time series data without any previous training. Such abilities are verified with extensive experiments.
\end{itemize}

\begin{figure*}[ht]
 \centering
 \includegraphics[trim={1.8cm 3.9cm 2.3cm 6cm},clip,width =.925\linewidth]{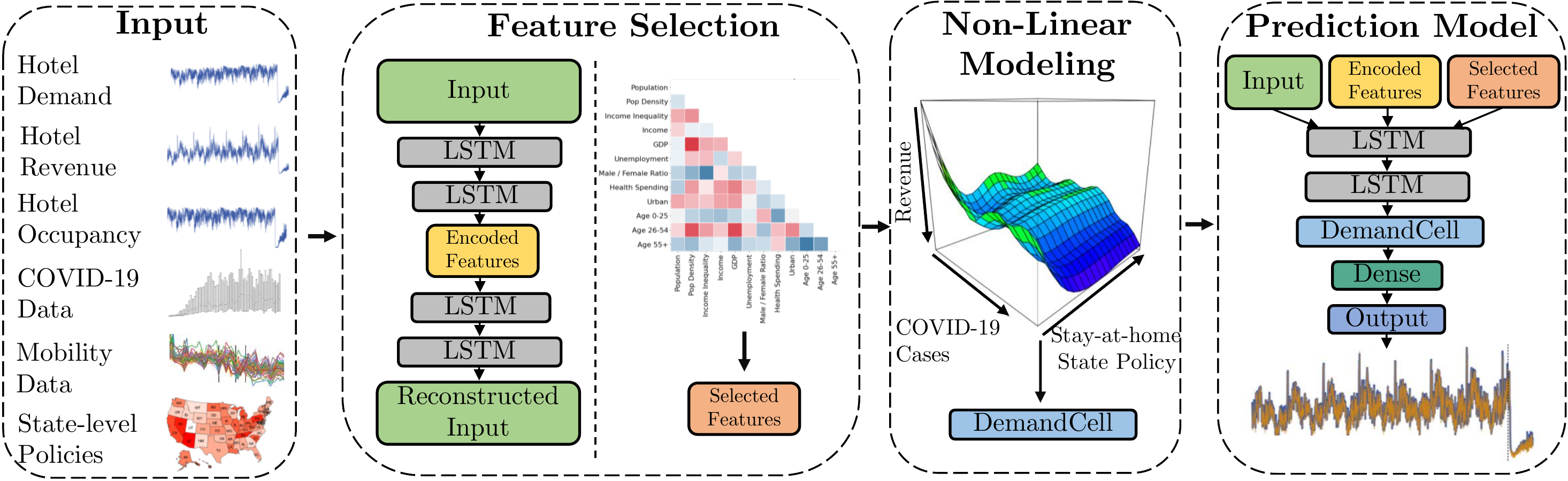}
 \centering
 \caption{DemandNet consists of three main components: (1) the feature selection mechanism which provides the top static and dynamic features; (2) a nonlinear model which provides interpretability and captures the effect of the selected features; (3) a prediction model which reports the estimated forecast.}
 \label{fig:framework}
\end{figure*}

\noindent\textbf{Organization.} The rest of the paper is organized as
follows. Section~\ref{section-preliminary} formulates the demand prediction problem.
Section~\ref{section-methodology} proposes DemandNet as a prediction method with dropout optimization for ensuring reliable uncertainty. Section~\ref{section-experiment} presents the datasets and metrics, discusses the experimental results, and conducts an ablation study. Section~\ref{section-related} summarizes the related works and Section~\ref{section-conclusion} concludes this paper.
\section{Problem Formulation}
\label{section-preliminary}

In this work, we are interested in the task of time series prediction under the presence of COVID-19 pandemic. Mathematically, given a dataset with $K$ univariate time series $\mathbf{D} = \{\mathbf{x}^{(1)},\mathbf{x}^{(2)},\ldots,\mathbf{x}^{(K)}\}$, we use
$\mathbf{x}^{(k)}=\{x^{(k)}_{1}, x^{(k)}_{2},\ldots, x^{(k)}_{T}\}$ to denote a period of data with length $T$ where
$\mathbf{x}^{(k)} \in \mathbb{R}^{T}$.

For every observation in $\mathbf{x}^{(k)}$, the corresponding covariates are denoted as $\mathbf{V}^{(k)} = \{\mathbf{v}^{(k)}_1, \mathbf{v}^{(k)}_2,\ldots, \mathbf{v}^{(k)}_{T}\}$. The covariates include up to $M$ exogenous variables (e.g., COVID-19 cases, mobility), where $\mathbf{v}^{(k)}_{t} \in \mathbb{R}^{M}$. As an example, the level of state policy at time $t$, can be shown as $\mathbf{v}^{(k)}_{t} \in \mathbb{R}$; otherwise, $\mathbf{v}^{(k)}_{t} = 0$ indicates that there are no state policy at time $t$. Hence, given the previous $\tau$ observations $\widehat{\mathbf{X}}_{t-\tau+1:t}^{(k)} = \{({\mathbf{x}}^{(k)}_{t-\tau+1},\mathbf{v}^{(k)}_{t-\tau+1}), ({\mathbf{x}}^{(k)}_{t-\tau+2},\mathbf{v}^{(k)}_{t-\tau+2}),\ldots, ({\mathbf{x}}^{(k)}_{t},\mathbf{v}^{(k)}_{t})\}$, the objective is predict till the $H$-th horizon time step $\mathbf{Y} = \mathbf{x}^{(k)}_{t+1:t+H}$. To this end, the optimization problem of DemandNet is defined as below:
\begin{equation}
    \mathbf{\Phi}^{*} = \text{argmin}_{\mathbf{\Phi}} \: \mathcal{L}\left(\mathcal{F}_\mathbf{\Phi}(\widehat{\mathbf{X}}),\mathbf{Y}\right)
\end{equation}
where $\mathbf{\Phi}$ are the parameters of the non-linear function $\mathcal{F}(\mathbf{X})$ and $\mathcal{L}$ is the loss function.

\section{DemandNet Framework}
\label{section-methodology}

We propose the DemandNet framework to model and predict the effect of the COVID-19 pandemic on the target times series as shown in Figure \ref{fig:framework}. There are three main components in DemandNet: First, DemandNet uses a feature selection mechanism to select the top correlated features (Section \ref{subsec:featureSelection}). Second, the nonlinear modeling of hotel sales is derived based on anti-contagion policy (Section \ref{subsec:Modelingsection}). Third, the prediction model uses the selected features and the derived nonlinear model to provide robust and reliable multi-step predictions (Section \ref{subsec:LSTM}).

\subsection{Feature Selection}
\label{subsec:featureSelection}
Many frameworks designed for pandemic time series data do not incorporate top important features about their subjects~\cite{timesspyros}. Research shows that a set of important features can improve the learning process \cite{introductionreliabilityintimeseries,tmisbenchmarkanalysis}. However, a great number of features can lead to the curse of dimensionality, which limits the model's ability to incorporate essential features. To resolve this issue, we develop a feature selection mechanism that retains the important features.

With the current advances in feature extraction strategies, deriving a complete set of information from a pandemic and its impact on hotel demand and revenue is challenging. Often the relationship between the
targeted variable and exogenous variables is unknown. It also is difficult to assume that all information is essential and relevant \cite{lim2019temporal}. Therefore, we need a specific design for feature selection that filters the appropriate related information. To achieve this objective, we develop a two-step feature selection mechanism to select each time series's top static and dynamic features.
The detailed procedure includes two steps:

{\bf First}, our feature selection mechanism adopts Spearman rank correlation to filter the non-monotonic static features. Compared with Pearson correlation that evaluates the linear relationship of features, Spearman rank correlation provides the monotonic relation of features that tend to change together but not at a constant (i.e., linear) rate. 

We start by deriving the correlation between every two sets of static features $a_{1:n}$ and $b_{1:n}$, each with $n$ number of observations. Both pairs of features will then be ranked in ascending order based on their raw value. For identical values, we assign a rank equal to the average of their positions in the ascending order. To this end, the correlation coefficient $r_{\mathrm{S}}$ of each pair can be calculated as:
\begin{equation}
r_{\mathrm{S}}=\frac{\sum_{i=1}^{n}\left\{\left(a_{i}-\bar{a}\right)\cdot\left(b_{i}-\bar{b}\right)\right\}}{\sqrt{\sum_{i=1}^{n}\left(a_{i}-\bar{a}\right)^{2}} 
\cdot
\sqrt{\sum_{i=1}^{n}\left(b_{i}-\bar{b}\right)^{2}}}
,\end{equation}
where both $\bar{a}$ and $\bar{b}$ are the average ranks. The coefficient can yield a value $-1 \leq r_{\mathrm{S}} \leq 1$, which can be interpreted as having a minimum association ($r_{\mathrm{S}}$ = 0) or as a perfect monotonic relationship ($r_{\mathrm{S}}$ = $\pm1$).Then, the feature selection mechanism removes the weak to moderate correlated features based on a region suggested by \cite{methoCorreCoef} (i.e., $- 0.3 < r_{\mathrm{S}} < 0.3$).

\begin{figure}[t]
 \centering
 \includegraphics[width =1.05\linewidth]{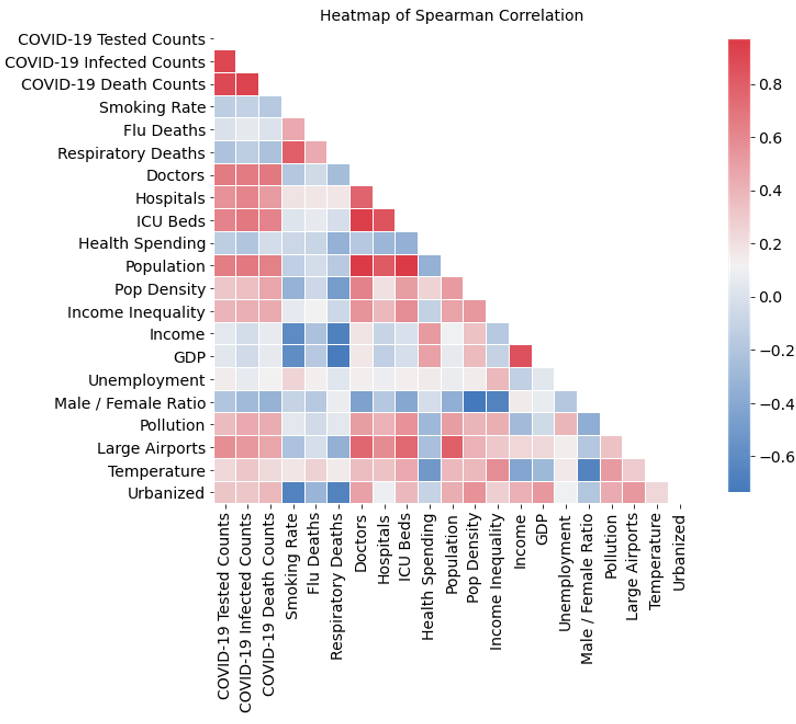}
 \caption{Heatmap of Spearman rank correlation for state-level static features. The feature selection mechanism removes the weakly correlated features.}
 \label{fig:heatmap}
\end{figure}

The exemplary Spearman Rank Correlations are shown in Figure \ref{fig:heatmap}, where it illustrates a heatmap of correlations from a set of static state-specific data. Such a correlational heatmap can be used to develop a profile for each time series and further improve the prediction accuracy as the model has access to important state-level economic, social, and health data \cite{lim2019temporal}. Moreover, such correlations enable DemandNet to use highly correlated features for developing profiles for location-specific time series data. To provide statistical inference, entropy-based correlations can also be a good candidate if we are given a greater sample size. 

{\bf Second}, complex time series data often contain critical features during the presence of external shocks that are hidden even by the most advanced methods. To resolve this issue, we stack Long-Short Term Memory (LSTM) layers to form serve as a Stacked Autoencoder. 

Specifically, the LSTM layers encode the time series into its top essential dynamic features (Figure \ref{fig:framework}). The input signal is reconstructed at the output through an intermediate layer with a reduced amount of neurons. The middle (smallest) layer is chosen as a dense layer to hold the essential features needed for a proper reconstruction. Intuitively, the model retains the deep and abstract features in the dense layer so that the reconstruction would be as close as possible to the original input. 
Note that we stack LSTM layers to capture complex abstract information that is difficult to gather by shallower layers.

The encoder maps the input $\hat{ \mathbf{x}} \in \mathbb{R}^{\mathrm{d}}$ to the encoded features $\mathbf{u} \in \mathbb{R}^{\mathrm{h}}$ in the encoding step where $\mathrm{h}$ is the number of neurons in the encoded features layer. Ideally, the number of neurons in the encoded feature ($\mathbf{u}$) should be chosen smaller than the stacked LSTM layers, so the network is compelled to keep only the top important abstract information. To this end, the decoder maps the encoded features to a reconstructed $\mathbf{z} \in \mathbb{R}^{\mathrm{d}}$ in the decoding step. 

We initialize the weights of the encoder and decoder randomly. Mathematically, the encoder $\mathcal{B}$ and decoder $\mathcal{D}$ are defined as:
\begin{equation}
 \mathbf{u}=\mathcal{B}\left(\mathbf{w}_{u} 
 \hat{\mathbf{x}}+b_{u}\right)
\end{equation}
\begin{equation}
 \mathbf{z}=\mathcal{D}\left(\mathbf{w}_{z} \mathbf{u}+b_{z}\right)
\end{equation}
where $\mathbf{w}_{u}$ and $\mathbf{w}_{z}$ denote the encoder and decoder weights. Coupled with $b_{u},$ and $b_{z}$ that denote the bias of hidden and output units. 

Then, we train the encoder-decoder network to derive the optimized parameters so that the error between $\hat{ \mathbf{x}}$ and $\mathbf{z}$ gets minimized, i.e.,

\begin{equation}
 \underset{\mathbf{w}_{x}, \mathbf{w}_{z}, b_{x}, b_{z}}{\text{argmin}} \operatorname{error}(\hat{ \mathbf{x}}, \mathbf{z}).
 \label{eq:error}
\end{equation}

The training process is an iterative update of the parameters $w$ and $b$, an update by which the error between the input e series and the reconstructed one at the network's output is steadily reduced until it is below a preset threshold. An effective training results in a decreased error, as displayed in Eq. \eqref{eq:error}, which ensures the proper internal features are present as the encoded features $\mathbf{u}$.

Thus, the two-step feature selection procedure provides DemandNet with top features that can be used in the nonlinear model and multi-step time series prediction methods discussed in the next sections.

\subsection{Nonlinear Modeling}
\label{subsec:Modelingsection}

Based on the top features selected in the previous section, we can now develop nonlinear models to discover the correlation of these features to the target data. Specifically, we are using a multilayer neural network to serve as a nonlinear model. Other than the interpretability, the nonlinear model can then be used as a control mechanism to improve the robustness of multi-step forecasts.

The multilayer NN is composed of the input ($\hat{ \mathbf{x}}$), hidden layer ($\mathbf{h}$), where the hidden layer $j$ can be represented as follows: 
\begin{equation}
\mathbf{h}_j=\mathcal{G}(\mathbf{w}_j \hat{ \mathbf{x}}+b_j),\end{equation}
where $\mathbf{w}_j, b_j, \mathbf{h}_j$ denote the weight, bias, and output of the $j$th hidden layer respectfully. Moreover, $\mathcal{G}$ denotes a nonlinear function (i.e., sigmoid). 

The modeling process starts by assigning random weights and biases for the hidden layers. Next, we feed the network our normalized data through the input layer, which then gets delivered to the hidden layers. Note that the network should be designed to avoid extreme values of weights to optimize the fit. Failure to do so will result in predictions that will be unstable for extrapolation. 
Hence, we use a penalty function to obtain a more stable and smooth fit.
To this end, the weights and biases of the hidden layers are chosen to minimize the following loss function:
\begin{equation}
L(y,y^{*},\mathbf{w})=\frac{1}{n} \sum_{i=1}^{n}(y_{i}-\hat{y_{i}})^{2} +\lambda \sum_{j=1}^{m} \mathbf{w}_{j}^{2}
,\end{equation}
where $y$ is the predicted output, $y^{*}$ is the original output, and $\lambda$ is used as the weight decay so that the weights would not hold extreme values. Consequently, the weights of the network are updated at every $e$ iteration by the following: 

\begin{equation}
\mathbf{w}_{j(e)} \longleftarrow \mathbf{w}_{j(e-1)}-\eta \frac{\partial L(y,y^{*},\mathbf{w})}{\partial \mathbf{w}_j}
,\end{equation}
where $\eta$ is the network's learning rate and is used to control the rate of changes to weights. Note that the updated network weights are hard to be interpretable just by their raw values. However, we can provide interpretability by observing the effect of the features on the target. This can be done by capturing the estimated target based on the marginal changes on one or more features as other features are held fixed. Upon construction of the nonlinear model, we can estimate the parameters of such model through the estimation of polynomial coefficients. We advise the use of such estimation to further reduce the computation time. Consequently, such nonlinear relation between target and features provides interpretability on the role of features which can then be leveraged in the forecast model.

\begin{figure}[!t]
 \centering
 \includegraphics[width=\linewidth]{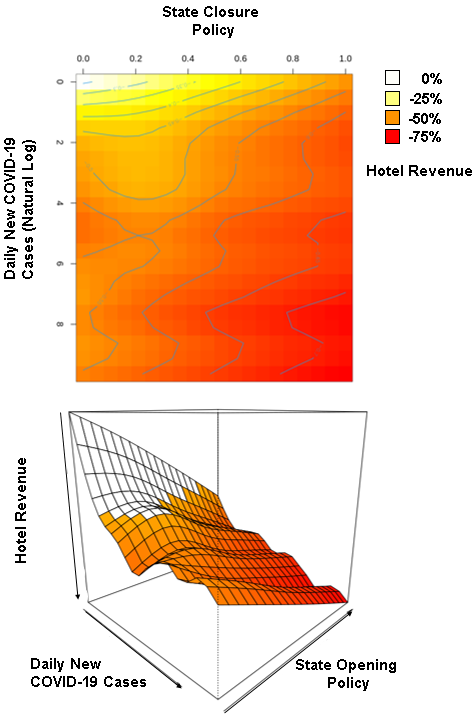}
 
 \caption{Nonlinear model of hotel revenue by the natural log of daily new cases and closure policy for the state of Florida. The closure policy changes from open (0) to the closure of all counties in a state (1).}
 \label{fig:model}
\end{figure}

As an example, we show the effect of each feature individually while keeping the other predictors fixed at their mean values. As seen in Figure \ref{fig:model}, the effect of features is reported by keeping the other features at their mean value while slightly changing the predictor of the nonlinear model. What stands out in Figure \ref{fig:model} is the nonlinear model's ability to derive the nonlinear relation of multiple features to the target (e.g., hotel revenue). Moreover, one of the key nonlinear relations is the state's closure and opening policies over time for individuals and businesses. Such nonlinear relation of policy holds great importance in multi-step forecasts where the DemandNet needs to apply the policies in a dynamic manner for the task of multi-step prediction. To this end, DemandNet is able to store such nonlinear relations and leverage them for future predictions.

\subsection{Prediction Model}
\label{subsec:LSTM}

\begin{figure}[!t]
 \centering
\includegraphics[trim={7cm 2.1cm 5.5cm 1cm},clip,width =\linewidth]{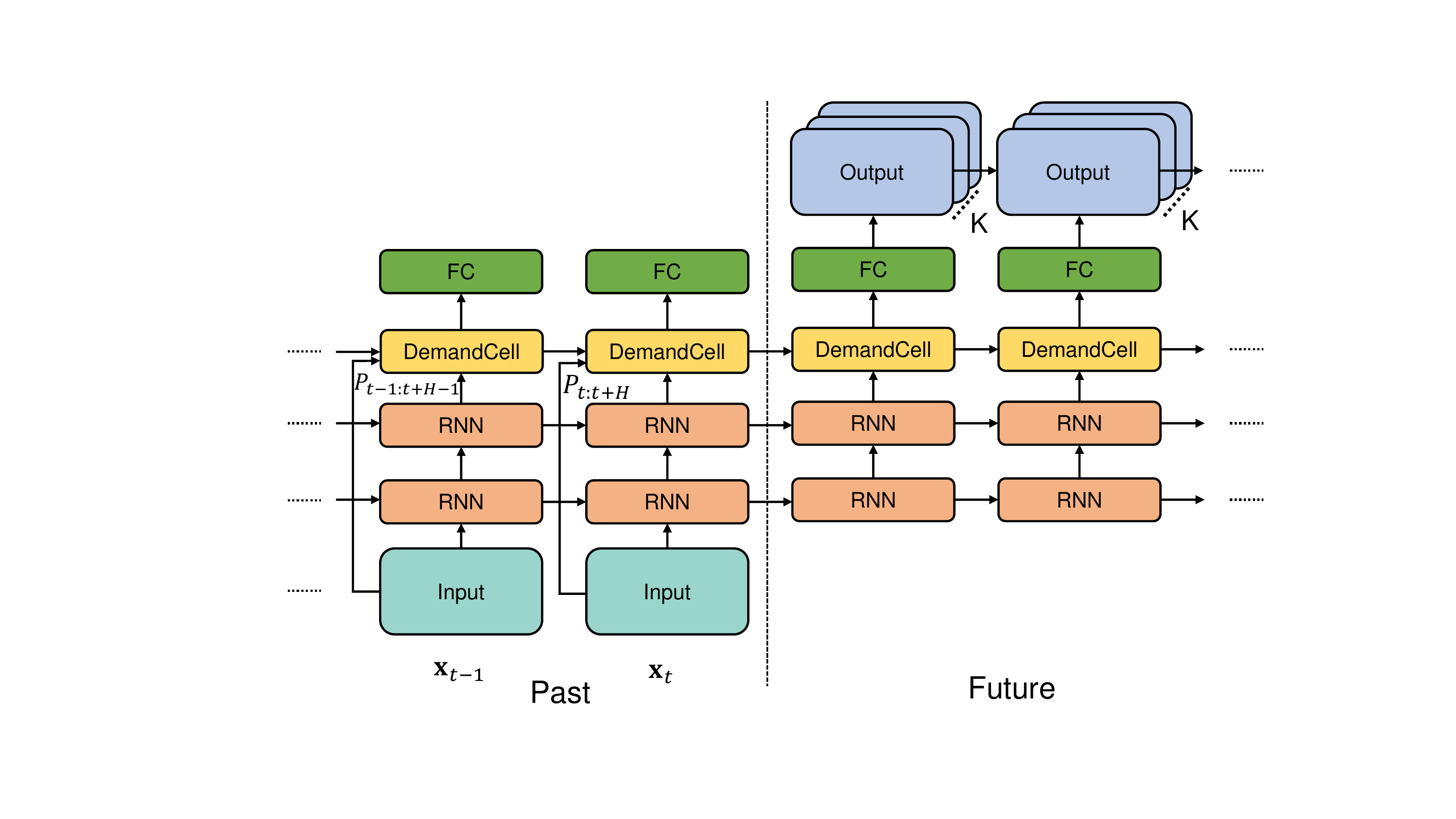}
 \hfill

 \caption{Multi-step time series prediction model. DemandCell is developed on top of existing RNNs (LSTMs or GRUs) to control the vector of dynamic state policies $P_t$ and provide confidence in its final prediction $(\bar{y}_{t})$.}

 \label{fig:rnn}
\end{figure}

The two main outcomes of the previous components are the top correlated features of data and the nonlinear relation of policies with respect to the prediction targets. Thus, such information can be leveraged for the task of multi-step time series prediction. Particularly, DemandNet feeds the selected features into its prediction model and uses the state's nonlinear relation with the prediction target in its control cell, namely, DemandCell. As shown in Figure~\ref{fig:rnn}, DemandCell is employed on top of well-known recurrent cells like LSTM. Specifically, it is designed to control the effect of state anti-contagion policies for the task of multi-step prediction, where the impact of such policies could have a long-lasting effect on consumer spending. 
The detailed procedure for the multi-step time series prediction is described below:


Given the input $\hat{ \mathbf{x}}$, the multi-step time series prediction model is denoted as $y = \mathcal{F}_{W}(\mathbf{x}),$ where $\mathbf{x}$, $y$, and $W$ stand for input data, output, and model weights. As shown in Figure \ref{fig:rnn}, the model $\mathcal{F}_W$ can consists of any RNNs that deals with vanishing gradient problem such as LSTMs \cite{lstmog} or Gated Recurrent Networks (GRUs) \cite{gruog}.
GRUs are similar to LSTMs in a manner that the forget and input gates of LSTM get combined into a single ``update'' of GRUs gate. Additionally, the cell and hidden state are combined into the ``reset'' gate, which makes GRUs more efficient than LSTMs in terms of complexity. 

We then employ DemandCell on top of the mentioned RNNs to control the dynamic anti-contagion policies over time. As shown in \ref{fig:rnn}, there is a skip connection between the input and DemandCell, where the information of the state's policy is used to control the output of the network. At the time step $t$ where the prediction is required, the framework uses a skip connection to feed DemandCell with $P_{t:t+H}$, which includes a vector of variables containing information about policies for $H$ days ahead. Note that DemandCell leverages the nonlinear model derived from the previous section to estimate the output accordingly based on the newly fed policies. Among the anti-contagion policies, we leverage the derived nonlinear relation of closure or reopening policy. Note that for predicting an unseen time series with unknown policies, such policies can be applied as a dummy variable to leverage the previously derived effects from every other state. 

Next, it is important to equip the model to report the level of uncertainty so that the performance can be evaluated with trust for multi-step prediction. To this end, DemandNet uses the Monte Carlo Dropout mechanism to provide uncertainty in each of its predictions. The Monte Carlo Dropout is implemented after every recurrent (LSTM or GRU) layer, i.e., randomly dropping out each hidden unit with a certain probability of $p$. The probability of dropout varies from zero to one and determines the likelihood of connection from the hidden layer between RNN layers. The probability of this dropout is dynamically optimized for each prediction, and as the dropout is applied at random, each predicted value would differ from each other. 
Specifically, the prediction of the original target data $y^*$ is done with the previously trained $\hat{\mathcal{F}}_{W}$, it generates $\mathcal{K}$ outputs $y$ to obtain the prediction distribution $\left\{y_{(1)}, \ldots, y_{(\mathcal{K})}\right\}$. By taking the mean of the derived distribution, we have:

\begin{equation}
 \bar{y}=\frac{1}{\mathcal{K}} \sum_{k=1}^{\mathcal{K}} {y}_{(k)}
\end{equation}

Then, the variance of the prediction distribution quantifies the prediction confidence:

\begin{equation}
{\operatorname{Var}}\left(\hat{y}\right)=\frac{1}{\mathcal{K}} \sum_{k=1}^{\mathcal{K}}\left({y}_{(k)}-{y}^{*}\right)^{2}
\end{equation}

To that end, we can use the mean of the distribution as a point prediction and its variance to report such prediction's confidence.

\section{Experiments}
\label{section-experiment}
This section reports the conducted experiments on DemandNet. Specifically, we provide a summary of the datasets in Section \ref{subsec:dataset}. Then, the baseline methods and evaluation metrics are reported in Sections \ref{subsec:methodsofcomparison} and \ref{subsec:metrics} respectfully. We then specify experimental settings in Section \ref{subsec:settings}. Finally, we report the results of the experiments and discuss their interpretation in Section \ref{sec:results}.

\begin{table*}[ht]
 \centering
 \caption{Description of Datasets Used in Experiments.}
\begin{tabular}{c|ccccc}
\toprule 
Dataset & Observations & Min & Max & Mean & SD\\
\midrule
\rowcolor{black!20} Hotel Revenue & 25803 (Daily) & 88.9K \$ &4.34M \$ & 1.86M \$ & 691K \$ \\ 
Hotel Demand & 25803 (Daily) & 1174 &26927 & 5156 &18799\\ 
\rowcolor{black!20} Hotel Occupancy & 25803 (Daily)&7.57\% & 99.09\%&18.44\%&74.01\%
\\
\bottomrule
\end{tabular}
 \label{tab:dataset}
\end{table*}
\subsection{Datasets}
\label{subsec:dataset}
In this experiment, we are selecting real-world time series datasets\footnote{Datasets are available at
https://github.com/ashfarhangi/COVID-19}. 
Our main dataset contains the hotel demand and revenue of 8 major tourist destinations in the US (e.g., Orlando, Los Angeles, ...). The dataset is provided by~\cite{str} and contains daily occupancy, demand, and revenue of the upper-middle class hotels. 

The second dataset contains the daily report of new COVID-19 cases, deaths gathered by Johns Hopkins University~\cite{COVID-19hopkins}.

The third dataset contains the mobility dataset gathered by Google~\cite{aktay2020google}, which provides the relative changes of the resident's mobility prior to the pandemic (January 2020) in high-risk environments such as public transport and workplaces.

We also gathered dynamic exogenous variables such as the state's closure/open policy to enrich our dataset. Specifically, we gathered various static features such as the number of hospitals, GPD, and population, which was previously shown in Figure \ref{fig:heatmap}.

\subsection{Methods for Comparison}
\label{subsec:methodsofcomparison}
The baseline methods for comparison include:
\begin{itemize}
 \item Exponential Smoothing \cite{experimentEShydman}: A traditional time series prediction method that uses an exponential window function to provide a smooth prediction.
 \item ARIMA~\cite{ArimaOG}: A classical autoregressive integrated moving average method for time series prediction and is often used as a baseline.
 \item AE-LSTM~\cite{aelstm}: An LSTM network that uses autoencoder for deep feature extraction and provides a deterministic prediction. 
 \item Extreme-Event Forecasting~\cite{uber0}: A time series prediction model that uses Monte Carlo Dropout to extract the deep features of time series and provides uncertainty prediction.
 \item MQ-Transformer~\cite{stateartmqtransformer}: A transformer-based deep learning model that benefits from static features and future predefined labels. It also provides a state-of-art prediction in multiple datasets. 
\end{itemize}

\subsection{Metrics}
\label{subsec:metrics}
For providing a comprehensive evaluation, we adopt three different evaluation metrics. 
The first two evaluation metrics are mean absolute error (MAE) and root mean square error (RMSE) that are scale-dependant measures. MAE is denoted as $\mathbf{M A E}=\frac{1}{N} \sum_{i=1}^{N}\left|y_{i}-\hat{y}_{i}^{*}\right|,$
and RMSE is defined as
$\mathbf{R M S E}=\sqrt{\frac{1}{N} \sum_{i=1}^{N}\left(y_{i}-{y}_{t}^{*}\right)^{2}}$ where $y_{i}$ is the predicted value and ${y}_{i}^{*}$ is the original value. The third evaluation metric is the standard deviation (SD) that is correlated to the confidence of prediction and is denoted as $\mathbf{S D} = \sqrt{\frac{1}{N} \sum_{i=1}^{N}\left(y_{i}^{}-\bar{y}\right)^{2}}$ where $\bar{y}$ is the mean of the predicted distribution.

\begin{table}[!t]
 \begin{center}
 \caption{Hyperparameters of DemandNet.}
 \end{center}
 \centering
\begin{tabular}{l|r}
\hline Parameter & Values\\
\toprule
\rowcolor{black!20} Batch Size & 128 \\
Learning Rate & $1 \times 10^{-5}$ \\
\rowcolor{black!20} Weight Decay & $1 \times 10^{-6}$ \\
Multilayer NN Layers & 2\\
\rowcolor{black!20} RNN Hidden Units & 128\\
 RNN layers & 2\\
\rowcolor{black!20} Epochs & 100 \\
 Training Time & 10m \\
\bottomrule
\end{tabular} 
\label{tab_hyper}
\end{table}

\begin{table*}[!ht]
\centering
\caption{Performance comparison of time series prediction methods.
}
\begin{tabular}{ll|cccc}
& & \multicolumn{4}{c} {Forecast Horizon (Days)} \\
\hline Methods & Metrics & 10 & 20 & 40 & 80\\
\hline Exponential Smoothing & MAE & 0.09423 & 0.12412 & 0.23631 & 0.53242 \\
\cite{introhydmanExpSmooth} & RMSE & 0.12324 & 0.15634 & 0.25341& 0.55634 \\

\hline ARIMA & MAE & 0.05853 & 0.11487 & 0.23873 & 0.45387
 \\
\cite{hyndman2013coherent}& RMSE & 0.07103 & 0.13268& 0.25585 & 0.48695\\
\hline EE-Forecasting & MAE & 0.04548 & 0.04695 & 0.05987 & 0.06122 \\
\cite{uber0}& RMSE & 0.04878 & 0.04921 & 0.06548 & 0.06587 \\
 & SD & 0.00439 & 0.00527 & 0.00648 & 0.00631 \\
 \hline MQ-Transformer & MAE & 0.02210 & 0.02261 & 0.03352 & 0.03731 \\
 \cite{stateartmqtransformer}& RMSE & 0.02831 & 0.03724 & 0.04112 & 0.04312 \\
 & SD & 0.00450 & 0.00551 & 0.00553 & 0.00605
 \\
\hline DemandNet-LSTM & MAE & \textbf{0.01257} & 0.01655 & 0.01852 &0.02263 \\
(Ours) & RMSE & \textbf{0.01442} & 0.01831& 0.02178 & 0.02434 \\
 & SD & \textbf{0.00356}& 0.00351 & 0.00461 & 0.00612 \\ 
\hline DemandNet-GRU & MAE & 0.01302 &\textbf{0.01357} & \textbf{0.01424} & \textbf{0.02048} \\
(Ours)& RMSE & 0.01504&\textbf{0.01512}&\textbf{0.01734}&\textbf{0.02343} \\
 & SD & 0.00398& \textbf{0.00348} & \textbf{0.00451} & \textbf{0.00568} \\
\hline
\end{tabular}
\label{tab:table8020}
\end{table*}

\subsection{Experimental Settings}
\label{subsec:settings}
The models are implemented using Python $3.7$ and tested on a cloud workstation that contains four Intel Xeon 2.3GHz CPUs, 32GB RAM, and one Nvidia P100 GPU. We conducted a grid search over all tunable hyperparameters on the held-out validation set for baseline methods and our proposed method and reported the results (Table 2). 

From the baseline methods, we applied the widely used Exponential Smoothing and ARIMA by using Statsmodels library \cite{expStatsModel}. 
The nonlinear model and the prediction model are implemented by the TensorFlow~\cite{abadi2016tensorflow} library. The hyperparameters of all models are tuned properly to provide the least amount of error in their prediction accuracy. Specifically, we conducted a grid search over all tunable hyperparameters on a 10\% held-out validation set for the baseline methods and our framework. For both variations of DemandNet, the input window's search ranges are chosen from \{4, 8, 16, 32\}. Note that enlarging the time window for more than 32-time steps will result in a marginal improvement in accuracy but substantially increase the training time for a dataset with a high dimension of features. The dimensions of the multilayer NN and RNN hidden layers are chosen from $\{32,64,128,256\}$, and the number of layers is chosen from $\{1,2,4,8\}$. Note that enlarging such numbers would increase the computation time.

Among NN models, the batch size and training epoch are set to be 128 and 200, respectively. We used the same set of essential features extracted from the proposed feature selection mechanism for all baseline methods except for the ARIMA and Exponential Smoothing models, where they are not able to benefit from multidimensional features. 
Both variations of DemandNet can change the dropout probability dynamically for each prediction, and the SD is calculated based on $K=100$ predictions.

In this study, we propose two sets of experiments. The first experiment is a standard $80-20$ multi-step time series prediction evaluation method and is designed to evaluate the performance of DemandNet and baseline models. To this end, our dataset is divided into three subsets: training set (80\%), validation set (10\%), and the testing set (10\%). 
The second experiment is designed to test the performance of DemandNet exclusively on completely untrained time series data. This experiment is applicable to evaluate the framework's performance when a newly added set of time series data contains unknown features, or we need to abstain from the computational cost of training.

For both experiments, the performances of all models are not limited to one state or category, but the average observed losses for all test states and all six prediction categories. In this manner, our report is sufficient to justify the model's performance for the total US consumers.

All architecture hyperparameters are set in a fairground. We used the same input features for the baseline methods except for the ARIMA model, which does not benefit from the multidimensional features.

\subsection{Experimental Results and Discussions}
\label{sec:results}
To demonstrate DemandNet performance's consistency, we provide two comprehensive evaluations: the aforementioned $80-20$ testing (i.e., 80\% trained) and the testing on a set of entirely untrained data. In both cases, we calculated the MAE, RMSE, and SD and discussed their interpretation.

Table \ref{tab:table8020} showcases the performance of DemandNet to compare with baseline in 80-20 testing phases. Compared to the baseline model, DemandNet-GRU achieved the highest accuracy for all horizon windows. However, the difference in its accuracy increases when the horizon forecast window increases.

Note that ARIMA and Exponential Smoothing models are designed to provide deterministic results \cite{introductionreliabilityintimeseries}. In contrast, EE-Forecasting and MQ-Transformer can provide the multi-step forecast with a level of confidence for each day. 
Among the baseline methods, the Exponential Smoothing and the ARIMA model's error increase significantly as the forecast horizon increases. This is because both models are not able to leverage the essential features and adapt properly to sudden changes during the pandemic. Moreover, both methods cannot employ the essential features provided by DemandNet because they can only employ the target time series. Among NN-based models, only DemandNet and EE-Forecasting \cite{uber0} benefit from the Monte Carlo Dropout mechanism. However, the EE-Forecasting is designed to use a static value of dropout probability through each experiment, while DemandNet adjusts this value dynamically for every forecast at each time step. Considering the selected features are the same for all these methods, we believe the lower SD is due to the dynamic dropout probability mechanism in DemandNet's prediction model. 
Overall, the two proposed models, DemandNet-LSTM and DemandNet-GRU, consistently outperform the baseline NN models \cite{stateartmqtransformer,uber0}, especially under longer horizon windows (40 and 80 days) due to their ability to incorporate state closure/open policy in their prediction. Considering all three evaluation metrics, DemandNet-GRU is best suited for the US consumer spending dataset due to reported higher confidence and accuracy. 

The second experiment is designed to evaluate the performance of DemandNet on unseen and untrained data exclusively. Testing on untrained data, as shown in Table \ref{tab:tableunseen}, holds importance when a new dataset is introduced to the network and might often hold unknown features. To this end, we kept away the complete data of four randomly chosen states in all three categories from the networks during the training phase and tested the performance of DemandNet on them. We believe that compared to 80-20 testing, such a testing procedure provides more significant insights into the model's robustness.

\begin{figure}[!t]
 \centering
 \includegraphics[width=\linewidth]{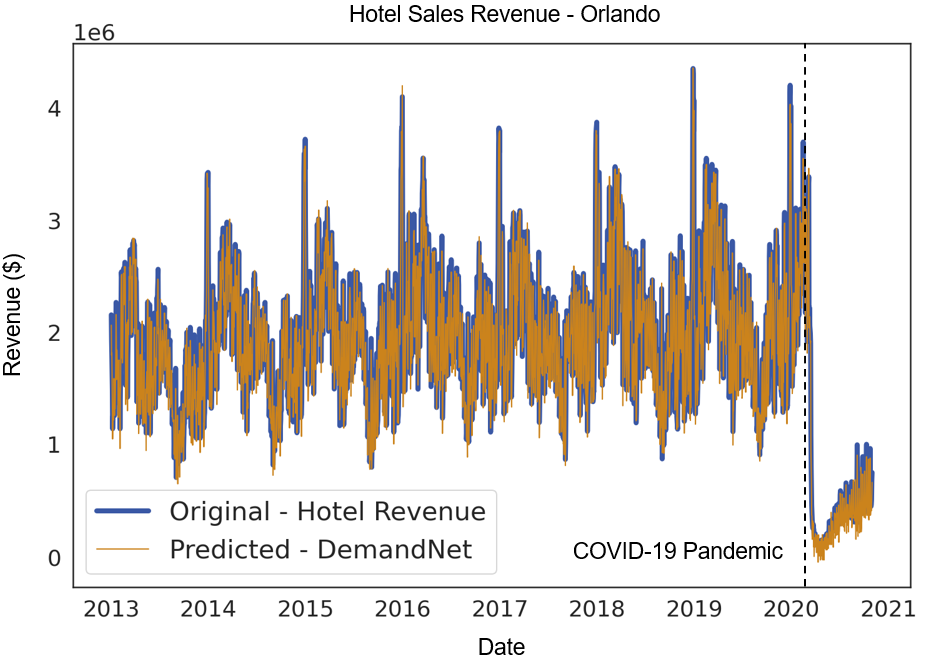}
 \caption{A sample forecast of DemandNet. Hotel sales of Orlando is the target. DemandCell aids in prediction of sudden drop during the COVID-19 pandemic.}
 \label{fig:unseen}
\end{figure}

Compared to 80-20 testing in Table \ref{tab:table8020}, the reported error for unseen data is higher and has less confidence in the forecasts for all forecast horizons. This is due to the model's inability to train on the kept away time series and not being able to follow the trend properly. Between the two variations of DemandNet, DemandNet-LSTM provides higher accuracy at smaller horizons (10 days) but falls behind DemandNet-GRU in longer horizons. However, we believe that overall, DemandNet-GRU is a more robust model as it handles longer horizons with greater accuracy and less uncertainty. 

\begin{table}[!h]
\caption{Performance of DemandNet-LSTM and DemandNet-GRU on completely unseen data.
}
\small
\hspace{-3mm}
\begin{tabular}{ll|cccc}
& & \multicolumn{4}{c} {Forecast Day Horizon} \\
\hline Method & Metrics & 10 & 20 & 40 & 80\\
\hline LSTM & MAE & \textbf{0.03464} & 0.03645 & 0.04341& 0.05653 \\
& RMSE & \textbf{0.03754} & 0.03883& 0.04524 & 0.05836 \\
& SD & \textbf{0.00531} & 0.00643 & 0.00736& 0.00833 \\
\hline
GRU & MAE & 0.03645 &\textbf{0.03345} & \textbf{0.04234} & \textbf{0.05533} \\
& RMSE & 0.03853&\textbf{0.03953}&\textbf{0.04475}&\textbf{0.05765} \\
 & SD & 0.00544 & \textbf{0.00653} & \textbf{0.00733} & \textbf{0.00856} \\
\hline
\end{tabular}
\label{tab:tableunseen}
\end{table}

Figure \ref{fig:unseen} demonstrates the prediction error of a typical case of a completely untrained time series dataset from one state (Florida). Both DemandNet-LSTM and DemandNet-GRU predictions follow the observed time series with great accuracy in the first days of the forecast horizon. However, the predictions' errors increase as the horizon windows increase comparably to the early days of the predictions. As shown in Figure \ref{fig:unseen}, the majority of loss occurs during the far horizon and is due to the framework's inability to predict the trend accurately. Compared to the 80-20 experiments, using the second procedure (unseen testing) can be an acceptable tradeoff when a newly added time series requires punctual prediction or exogenous features are unavailable.

\section{Related Works}
\label{section-related}

There have been a surge in the development of general neural network (NN) frameworks for investigating the impact of COVID-19 on time series business data~\cite{nemati2020machine1,nemati2020machine,relatedworksMLsurveyforCovid,nemati2020covid}. The majority of these studies leverage NN prediction models to handle the large-scale nature of pandemic data. 
To derive the nonlinear properties of such data, various nonlinear functions need to be fitted \cite{methnNNmodelingreview}. Such a process requires great effort and is typically avoided for large-scale time series data with dynamic properties. 

The authors in \cite{lim2019temporal} proposed a transformer framework that achieves a state-of-art model for multi-horizon prediction for various complex time series datasets. Their method leverages static covariates and known inputs to improve its performance. Furthermore, they employ a gating mechanism that allows their framework to handle large-scale datasets. Additionally, they leverage a temporal processing procedure that enables the network to capture the long-and short-term relationships of both observed and known inputs (e.g., days of a week). Moreover, their model employs a seq-seq layer for short-term and attention mechanisms for long-term dependencies. However, our work is different from~\cite{lim2019temporal} due to the feature selection mechanism to derive highly correlated static features so that the network can provide greater accuracy and confidence in the predictions. Another practical advantage of DemandNet is that it provides a more interpretable and simpler nonlinear prediction model, which is far less computationally expensive than the transformers. 

\section{Conclusion}
\label{section-conclusion}

We propose a novel deep learning framework, DemandNet, to provide a robust multi-step time series forecast amid the COVID-19 pandemic. A feature selection mechanism is proposed to estimate the correlation between the COVID-19 pandemic and time series input data. This deep learning framework provides valuable insight into the social, health, and economic static features that affect the market demand, business operations, and COVID-19 cases. A multilayer neural network is then designed to derive the nonlinear relation of the key features to the observed input. Our results from the nonlinear model show that the negative impact of business closure policy on hotel demand and revenue is higher than other location-specific exogenous variables (e.g., google mobility and COVID-19 daily new cases). Consequently, a novel RNN is developed to leverage the selected features and nonlinear model to provide a robust prediction with a high confidence level. We evaluate the performance of DemandNet based on hotel sales revenue, demand, and occupancy. Compared with the state-of-the-art baseline methods, DemandNet provides superior prediction accuracy with a higher confidence level. Our model can be expanded to model other business time series data under the influence of external shocks. Regarding future work, we plan to apply the DemandNet framework to understand the effect of various intervention policies on the COVID-19 transmissions in worldwide cities. 

\section*{Acknowledgment}
This work is supported by the US National Science Foundation under grants No. 1937833 and 2028481.

\bibliographystyle{ieeetr}
\bibliography{0.Main.bib}
\end{document}